\documentclass[10pt,twocolumn,letterpaper]{article}
\usepackage{authblk}
\usepackage[pagenumbers]{cvpr} 
\usepackage{graphicx}
\usepackage{amsmath}
\usepackage{amssymb}
\usepackage{booktabs}
\usepackage{float}
\usepackage{xcolor}
\usepackage{mathtools, nccmath}

\usepackage{soul}
\usepackage{xspace}\xspace
\usepackage{array}
\usepackage{dblfloatfix}
\usepackage{transparent}
\usepackage{adjustbox}
\usepackage{algorithm}
\usepackage{listings}

\usepackage[pagebackref,breaklinks,colorlinks]{hyperref}

\usepackage[capitalize]{cleveref}
\crefname{section}{Sec.}{Secs.}
\Crefname{section}{Section}{Sections}
\Crefname{table}{Table}{Tables}
\crefname{table}{Tab.}{Tabs.}

\def\confName{CVPR}
\def\confYear{2023}

\begin{document}

\title{Decompose, Adjust, Compose: Effective Normalization by Playing with Frequency for Domain Generalization}

\author[1]{Sangrok Lee\thanks{Equal contribution}}
\newcommand\CoAuthorMark{\footnotemark[\arabic{footnote}]}
\author[2]{Jongseong Bae\protect\CoAuthorMark}
\author[1]{Ha Young Kim\thanks{Corresponding author}}

\affil[1]{Graduate School of Information, Yonsei University}
\affil[2]{Department of Artificial Intelligence, Yonsei University}
\affil[ ]{\{lsrock1, js.bae, hayoung.kim\}@yonsei.ac.kr}
\maketitle

\begin{abstract}
   Domain generalization (DG) is a principal task to evaluate the robustness of computer vision models. 
   Many previous studies have used normalization for DG.
   In normalization, statistics and normalized features are regarded as style and content, respectively.
   However, it has a content variation problem when removing style because the boundary between content and style is unclear. 
   This study addresses this problem from the frequency domain perspective, where amplitude and phase are considered as style and content, respectively.
   First, we verify the quantitative phase variation of normalization through the mathematical derivation of the Fourier transform formula. 
   Then, based on this, we propose a novel normalization method, $PCNorm$, which eliminates style only as the preserving content through spectral decomposition.
   Furthermore, we propose advanced $PCNorm$ variants, $CCNorm$ and $SCNorm$, which adjust the degrees of variations in content and style, respectively. 
   Thus, they can learn domain-agnostic representations for DG. 
   With the normalization methods, we propose ResNet-variant models, DAC-P and DAC-SC, which are robust to the domain gap. 
   The proposed models outperform other recent DG methods.
   The DAC-SC achieves an average state-of-the-art performance of 65.6\% on five datasets: PACS, VLCS, Office-Home, DomainNet, and TerraIncognita.

\end{abstract}

\section{Introduction}
Deep learning has performed remarkably well in various computer vision tasks.
However, the performance decreases when distribution-shifted test data are given\cite{QuioneroCandela2009DatasetSI}.
As training and testing datasets are assumed to be identically and independently distributed, common vision models are not as robust as the human vision system, which is not confused by affordable changes in the image style\cite{Hendrycks2019BenchmarkingNN}.
Domain generalization (DG) aims to learn models that are robust to the gap between source domain and unseen target domain to address this problem\cite{Wang2021GeneralizingTU}. Moreover, DG is challenging because models should learn domain-irrelevant representation in an unsupervised manner. 
\begin{figure}[t]
  \centering
   \includegraphics[width=1.0\linewidth]{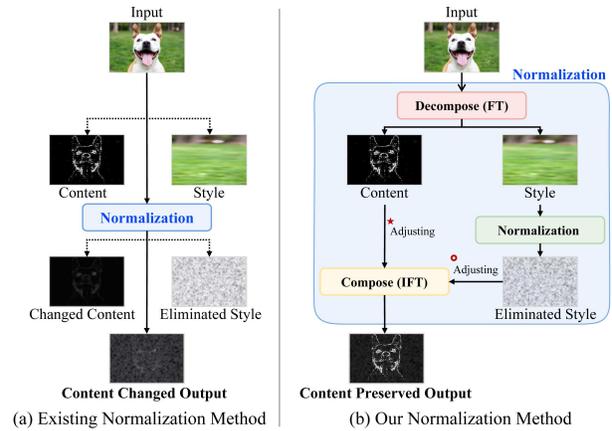}
   \caption{Concepts of (a) the existing normalization and (b) the proposed methods. Our methods prevent or adjust the content change caused by existing normalization using spectral decomposition.
   The solid line marks the feedforward process and the dashed line conceptually represents the content and style of the feature. Red-colored star and doughnut in (b) indicate the content and style adjusting terms, respectively.}
   \label{fig:onecol}
\end{figure}

The style-based approach is widely studied for DG,
which defines the domain gap as the difference in style\cite{Zhou2021DomainGW,Xu2021AFF,zhao2022test,Nam2021ReducingDG,Jin2020StyleNA}. 
Typically, normalization methods, such as batch normalization (BN)\cite{Ioffe2015BatchNA}, layer normalization (LN)\cite{Ioffe2015BatchNA}, and instance normalization (IN)\cite{Ulyanov2016InstanceNT}, which are well-known in style transfer, are used in this approach.
Normalization statistics contain style information, and normalization can successfully extract the style from a feature.
However, the content is also changed when the style is eliminated\cite{Jin2020StyleNA,Huang2017ArbitraryST,ibn}.

Moreover, another method for style-based DG\cite{Xu2021AFF,Wang2022DomainGV,Wang2020HighFrequencyCH,Chen2021AmplitudePhaseRR} is the frequency domain-based method. 
Input images are decomposed into amplitude and phase using the Fourier transform (FT)\cite{bracewell1986fourier}.
The amplitude and phase are each regarded as the style and content of the input image, respectively\cite{Piotrowski1982ADO,Chen2021AmplitudePhaseRR,Xu2021AFF,yang2020fda,oppenheim1981importance}.
Each component is manipulated independently to generate the style-transformed image.
In this context, this method has an advantage of the separation between style and content\cite{zhao2022test}.
Nevertheless, most previous studies have applied it to just the input-level data augmentation\cite{Chen2021AmplitudePhaseRR,Xu2021AFF,yang2020fda} for DG. 

\begin{figure}[t]
  \centering
   \includegraphics[width=0.6\linewidth,keepaspectratio]{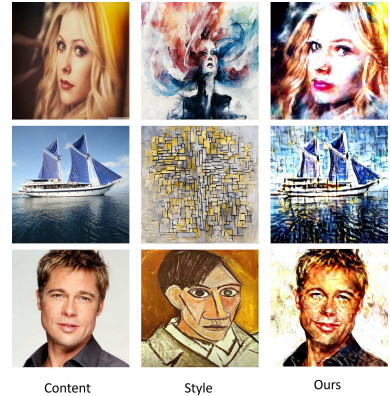}

   \caption{Examples of style transfer with spectral decomposition. Only the amplitude of the target images are transferred instead of their normalization statistics in AdaIN.}
   \label{fig:adain}
\end{figure}

Thus, the normalization is expected to be complemented by the frequency domain-based method if the method is also applicable at the feature level.
To identify the feasibility of this, we conduct a style transfer experiment. 
We replace IN in AdaIN\cite{Huang2017ArbitraryST}, a milestone work that uses normalization in style transfer, with spectral decomposition. 
The qualitative results in Fig. \ref{fig:adain} indicate that the frequency domain-based method can work as a feature-level style-content separator instead of normalization. 

Motivated by this, we aim to overcome the content change problem in normalization by combining the normalization with spectral decomposition.
The overall concept of our proposed method is visualized in Fig. \ref{fig:onecol}.
For this, we investigate the effect of the existing normalization in DG from the standpoint of the frequency domain.
We verify how normalization transforms the content of a feature by mathematically deriving the FT formula.
This is the first to present such an analysis.

Then, based upon the analysis, we introduce a novel normalization method, phase-consistent normalization ($PCNorm$), which preserves the content of a pre-normalized feature. The $PCNorm$ synthesizes a content-invariant normalized feature by composing the phase of pre-normalized feature and the amplitude of post-normalized feature. The experimental results reveal the effectiveness of $PCNorm$ in DG compared to existing normalization.

Along with the success of $PCNorm$, we take a step further and propose two advanced $PCNorm$ variants: content-controlling normalization ($CCNorm$) and style-controlling normalization ($SCNorm$). 
The main idea of both methods is not to preserve the content or style but to adjust the change in it.
$CCNorm$ and $SCNorm$ regulate the changes in content and style, respectively, so they can synthesize more robust representations of the domain gap.

With the proposed normalization methods, we propose ResNet\cite{resnet} variant models, DAC-P and DAC-SC. 
DAC-P is the initial model with $PCNorm$, and DAC-SC is the primary model using $CCNorm$ and $SCNorm$. 
In DAC-P, the existing BN in the downsample layer is replaced with $PCNorm$. 
In contrast, DAC-SC applies $CCNorm$ instead of $PCNorm$, and $SCNorm$ is inserted at the end of each stage.
We evaluate DAC-P and DAC-SC on five DG benchmarks: PACS, VLCS, Office-Home, DomainNet and TerraIncognita,
and DAC-P outperforms other recent DG methods with average performance of 65.1$\%$.
Furthermore, the primary model, DAC-SC, achieves state-of-the-art (SOTA) performance of 65.6$\%$ on average,
and displays the highest performance at 87.5$\%$, 70.3$\%$ and 44.9$\%$ on the PACS, Office-Home, and DomainNet benchmarks.

The contributions of this paper are as follows:
\begin{itemize}
\item For the first time, we analyze the quantitative shift in the phase caused by normalization using mathematical derivation.
\item We introduce a new normalization, $PCNorm$, which can remove style only through spectral decomposition.
\item We propose the advanced $PCNorm$ variants, $CCNorm$ and $SCNorm$, which can learn domain-agnostic features for DG by adjusting the degrees of the changes in content and style, respectively.
\item We propose ResNet-variant models, DAC-P and DAC-SC, which applies our proposed normalization methods. We experimentally show that our methods are effective for DG and achieve SOTA average performances on five benchmark datasets.

\end{itemize}
\label{sec:intro}

\section{Related Work}
\label{sec:related work}

\textbf{Domain-invariant learning for DG.} 
The main purpose of DG methods is to learn domain-invariant features. There are numerous approaches for DG.
Adversarial learning methods\cite{Ganin2016DomainAdversarialTO,Li2018DomainGW,fan2021adversarially,Li2018DeepDG,Matsuura2020DomainGU,Zhao2020DomainGV,Li2018DeepDG} prevent models from fitting domain-specific representations, allowing the extraction of domain-invariant representations only.  
Regularization methods introduce various regularization strategies, such as the reformulation of loss functions\cite{Li2022InvariantIB,sun2016deep}, gradient-based dropout\cite{Huang2020SelfChallengingIC}, and contrastive learning\cite{kim2021selfreg} for DG.
Optimization methods aim to reduce the distributional variance and domain-specific differences using kernel methods\cite{Blanchard2021DomainGB,Muandet2013DomainGV} and penalty on variance\cite{Krueger2021OutofDistributionGV}.
Meta-learning methods\cite{Li2018LearningTG,Zhang2021AdaptiveRM} simulate diverse domain changes for various splits of training and testing data from source datasets.

\textbf{Style-based learning for DG.}
Style-based learning methods define the domain gap as the style difference between each domain and try to extract style-invariant features.
There are two widely used methods that enable handling style and content of images, normalization and frequency domain-based methods.
Normalization\cite{segu2020batch,Zhou2021DomainGW,Nam2021ReducingDG} has been widely used to filter the style of features\cite{Huang2017ArbitraryST,Nam2018BatchInstanceNF,ioannou2022depth}.  
There have been various works for DG by applying normalization methods such as BN\cite{segu2020batch}, IN\cite{Zhou2021DomainGW,Nam2021ReducingDG}, and both of them\cite{pan2018two,fan2021adversarially}. 
Jin \etal \cite{Jin2020StyleNA} points out the downside of normalization that it causes the loss of content.

In frequency-based methods, the style and content of the image are represented by the amplitude and phase in frequency domain\cite{oppenheim1979phase,oppenheim1981importance,piotrowski1982demonstration,hansen2007structural}. 
Through FT, many works manipulate them to make models robust to the distribution shift. 
FDA\cite{yang2020fda} proposes the spectral transfer, which is similar to the style transfer, that the low-frequency component of source amplitude is transferred to that of the target amplitude. In \cite{Xu2021AFF}, the amplitude swap (AS) and amplitude mix (AM) strategies are introduced for data augmentation, the former is exactly the same as the spectral transfer\cite{yang2020fda} and the latter is to mix amplitude of source and target domain. 
Similarly, \cite{Wang2022DomainGV} implements data augmentation by applying multiplicative and additive Gaussian noises to both of the amplitude and phase of the source domain. 
Most frequency-based works have focused on input-level data augmentation. 
Different from those, we propose novel frequency domain-based normalization methods that are applicable at the feature level.

\begin{algorithm}[t]
\caption{Pseudocode of the Proposed Normalization Methods in PyTorch-like Style.}
\label{alg:code}
\definecolor{codeblue}{rgb}{0.25,0.5,0.5}
\lstset{
  backgroundcolor=\color{white},
  basicstyle=\fontsize{7.2pt}{7.2pt}\ttfamily\selectfont,
  columns=fullflexible,
  breaklines=true,
  captionpos=b,
  commentstyle=\fontsize{7.2pt}{7.2pt}\color{codeblue},
  keywordstyle=\fontsize{7.2pt}{7.2pt},
}



\begin{lstlisting}[language=python]
# decompose: frequency feature to amplitude and phase
# compose: amplitude and phase to frequency feature
# weight: learnable parameters, T_c, T_s: temperatures
def pcnorm(f): # f: a spatial feature
    f_norm = batch_norm(f) # batch_norm: BN
    F = FT(f) # FT: fourier transform
    F_norm = FT(f_norm)
    a, p = decompose(F)
    a_norm, p_norm = decompose(F_norm)
    f = IFT(compose(a_norm, p)) # IFT: inverse FT
    return f
    
def ccnorm(f):
    weight_c = softmax(weight/T_c, dim=0) 
    # mean: batch mean(train), cumulative mean(test)
    fc = f - mean * weight_c[0] 
    f_norm = batch_norm(f)
    FC = FT(fc)
    F_norm = FFT(f_norm)
    ac, pc = decompose(FC)
    a_norm, p_norm = decompose(F_norm)
    f = IFT(compose(a_norm, pc))
    return f

def scnorm(f):
    weight_s = softmax(weight/T_s, dim=0)
    f_norm = instance_norm(f) #instance_norm: IN
    F = FT(f)
    F_norm = FFT(f_norm)
    a, p = decompose(F)
    a_norm, p_norm = decompose(F_norm)
    f = IFT(compose(a_norm * weight_s[0] + a * weight_s[1], p)) 
    return f
\end{lstlisting}
\end{algorithm}
\section{Analysis}
\label{sec:analysis}
Prior to analysis, we clarify the terms of normalization.
We consider that normalization includes just BN, LN, and IN because these are all the existing normalization methods for DG, as far as we know.
As mentioned in Sec. \ref{sec:intro}, normalization has a content change problem. However, there is still no explicit verification of the problem.
Motivated by this, we examine it through a mathematical analysis from the frequency domain perspective. 
Specifically, we derive the quantitative change in the phase caused by normalization by expanding the FT formula.

\subsection{Spectral Decomposition}
We first explain the spectral decomposition\cite{castagna2006comparison,Piotrowski1982ADO}.
Generally, the term spectral decomposition has various meanings.
In this work, it refers to decomposing a feature into amplitude and phase using the discrete FT (DFT)\cite{sundararajan2001discrete}.

The DFT transforms the input spatial feature $f\in\mathbb{R}^{h\times w}$ into the corresponding frequency feature $\mathcal{F}\in\mathbb{C}^{h\times w}$ where $h$ and $w$ are the height and the width of $f$. $f(x,y)$ is an element of $f$ at image pixel $(x,y)$, and \(\mathcal{F}({u,v})\) represents the Fourier coefficient at frequency component \(\mathcal({u,v})\). We omit the round bracket term $(\cdot,\cdot)$ when we denote a feature composed of that element.
The DFT of $f$ is defined as follows:
\begin{equation}
\begin{split}&
\mathcal{F}(u,v) = \frac{1}{wh}\sum_{x=0}^{w-1}\sum_{y=0}^{h-1}{f(x,y)}\exp^{i2\pi(\frac{ux}{w}+\frac{vy}{h})} \\&
= \mathcal{F}_{real}(u,v) + i \,\mathcal{F}_{img}(u,v),
\end{split}
\label{eq:1}
\end{equation}
where $i$ is imaginary unit. In addition, $\mathcal{F}_{real}(u,v)$ and $\mathcal{F}_{img}(u,v)$ are the real and imaginary parts of $\mathcal{F}(u,v)$, respectively, as follows:
\begin{equation}
\resizebox{.9\columnwidth}{!}{$%
\begin{split}&
    \mathcal{F}_{real}(u,v) = \frac{1}{wh}\sum_{x=0}^{w-1}\sum_{y=0}^{h-1}{f(x,y)}\cos{2\pi(\frac{ux}{w}+\frac{vy}{h})},\\&
    \mathcal{F}_{img}(u,v) = \frac{1}{wh}\sum_{x=0}^{w-1}\sum_{y=0}^{h-1}{f(x,y)}\sin{2\pi(\frac{ux}{w}+\frac{vy}{h})}.
\end{split}
\label{eq:2}$%
}
\end{equation}

The process in which the spatial feature $f$ is decomposed into the amplitude $\alpha$ and phase $\rho$ is called spectral decomposition, where
$\alpha$ and $\rho$ are calculated as follows:
\begin{equation}
\begin{split}&
\alpha = \sqrt{\mathcal{F}_{real}^2 + \mathcal{F}_{img}^2},\\&
\rho = arctan{\frac{\mathcal{F}_{img}}{\mathcal{F}_{real}}}.
\end{split}
\label{eq:3}
\end{equation}
In addition, $\mathcal{F}$ can be reassembled from $\alpha$ and $\rho$:
\begin{equation}
\begin{split}&
\mathcal{F} = \alpha \, \cos(\rho) + i \,\alpha \, \sin(\rho).
\end{split}
\label{eq:4}
\end{equation}
We define a function of the disassembling frequency feature in the amplitude and phase as $decompose(\cdot)$ and the opposite process as $compose(\cdot)$: 
\begin{equation}
\begin{split}&
\alpha,\rho  = decompose(\mathcal{F}),\\&
\mathcal{F} = compose(\alpha,\rho).
\end{split}
\label{eq:5}
\end{equation}
\subsection{Content Variation by Normalization}
\label{sec:3.2}
In this section, we mathematically verify how normalization changes the content of the feature in the frequency domain.
We develop the FT formula of normalized feature $f^{norm}\in\mathbb{R}^{h\times w}$ and identify the relationship with that of the original feature $f$. 
Moreover, $f^{norm}$ is as follows:
\begin{equation}
\begin{split}&
{f}^{norm} = \frac{{f} - \mu}{\sigma},
\end{split}
\label{eq:6}
\end{equation}
where $\mu$ and $\sigma$ denote the statistical mean and standard deviation of ${f}$, respectively. Statistics are calculated differently as with normalization methods. In this study, however, $\mu$ and $\sigma$ are set as constants because they are computed in the same way in a channel.

Like Eq. \ref{eq:1}, the DFT of $f^{norm}$ is as follows:
\begin{equation}
\begin{split}&
\mathcal{F}^{norm}(u,v) = \mathcal{F}_{real}^{norm}(u,v) + i \,\mathcal{F}_{img}^{norm}(u,v).
\end{split}
\label{eq:7}
\end{equation}

In Eq. \ref{eq:2}, by the linearity property of FT, $\mathcal{F}_{real}^{norm}(u,v)$ and $\mathcal{F}_{img}^{norm}(u,v)$ are represented as follows:
\begin{equation}
\resizebox{.9\columnwidth}{!}{$%
\begin{split} &           
\mathcal{F}_{real}^{norm}(u,v) = \frac{1}{wh}\sum_{x=0}^{w-1}\sum_{y=0}^{h-1} 
\{\frac{f(x,y) - \mu}{\sigma}\} \cos2\pi(\frac{ux}{w}+\frac{vy}{h}),  \\& 
\mathcal{F}_{img}^{norm}(u,v) = \frac{1}{wh}\sum_{x=0}^{w-1}\sum_{y=0}^{h-1} \{\frac{f(x,y) - \mu}{\sigma}\} \sin2\pi(\frac{ux}{w}+\frac{vy}{h}).
\end{split}$%
}
\label{eq:8}
\end{equation}

Then we derive a relationship between $\mathcal{F}$ and $\mathcal{F}^{norm}$ by presenting Eq. \ref{eq:8} in terms of Eq. \ref{eq:2}:
\begin{equation}
\begin{split}&
    \mathcal{F}_{real}^{norm} = \frac{\mathcal{F}_{real}-\mathcal{F}_{real}^{\mu}}{{\sigma}},\\&
    \mathcal{F}_{img}^{norm} = \frac{\mathcal{F}_{img}-\mathcal{F}_{img}^{\mu}}{{\sigma}},
\end{split}
\label{eq:9}
\end{equation}
where $\mathcal{F}_{real}^{\mu}$ and $\mathcal{F}_{img}^{\mu}$ are real and imaginary parts of $\mathcal{F}^{\mu}\in\mathbb{C}^{h\times w}$. $\mathcal{F}^{\mu}$ is the frequency feature of $f^{\mu}\in\mathbb{R}^{h\times w}$, which is a feature whose elements are all ${\mu}$. 

In the same way as Eq. \ref{eq:3}, the amplitude and phase of $f^{norm}$, $\alpha^{norm}$ and $\rho^{norm}$, respectively, can be computed as follows:
\begin{equation}
\begin{split}&
    \alpha^{norm} = \frac{\sqrt{(\mathcal{F}_{real} - \mathcal{F}_{real}^{\mu})^{2} + (\mathcal{F}_{img} - \mathcal{F}_{img}^{\mu})^{2}}}{{\sigma}},\\&
    \rho^{norm} = arctan{\frac{\mathcal{F}_{img} - \mathcal{F}_{img}^{\mu}}{\mathcal{F}_{real} - \mathcal{F}_{real}^{\mu}}}.
\end{split}
\label{eq:10}
\end{equation}

By comparing $\rho^{norm}$ with $\rho$, we verify the numerical variation of the content information caused by normalization. 
We determine that $\rho^{norm}$, the phase of spatial feature $(f-\mu)/\sigma$, is same as the phase of $(f-\mu)$ because it is not affected by $\sigma$ (Eq. \ref{eq:10}). 
That is, the difference between $\rho^{norm}$ and $\rho$ in the frequency domain is simply caused by the mean shift in normalization in the spatial domain. Hence, we consider $\mu$ to be the content variation factor. As observed, the degree of content variation becomes greater when $\mu$ is larger.

\section{Proposed Method}
\label{sec:method}
In this section, based upon the analysis in Sec. \ref{sec:3.2}, we explain the novel normalization methods, $PCNorm$, $CCNorm$, and $SCNorm$, whose Pytorch-like pseudocode is described in Algorithm. \ref{alg:code}.
Next, we describe the ResNet\cite{resnet}-variant models: DAC-P and DAC-SC. 
\begin{figure}[t]
  \centering
   \includegraphics[width=1.0\linewidth]{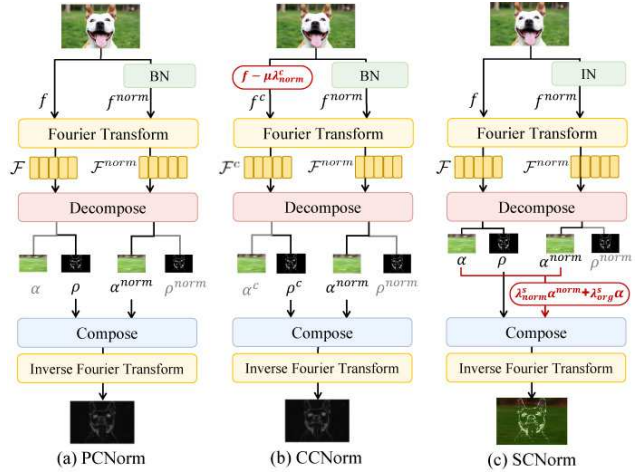}
   \caption{Illustrations of the proposed normalization methods. 
   Included notations are the same as in Sec. \ref{sec:analysis} and \ref{sec:method}. 
   \color{red}Red \color{black}  denotes adjusting operation in Sec. \ref{sec:ccnorm} and \ref{sec:scnorm}.
   }
   \label{fig:norms}
\end{figure}
\subsection{Phase Consistent Normalization (PCNorm)}
\label{sec:4.1}
In Sec. \ref{sec:3.2}, we verify that the difference between $\rho$ and $\rho^{norm}$ is caused by the mean shift by ${\mu}$ in normalization. Thus, we infer that a simple method to prevent content from changing is to avoid a mean shift in normalization.
We can circumvent content variation using the phase of pre-normalized feature, $\rho$, instead of $\rho^{norm}$.
\begin{figure*}
  \centering
    \includegraphics[width=1.\linewidth]{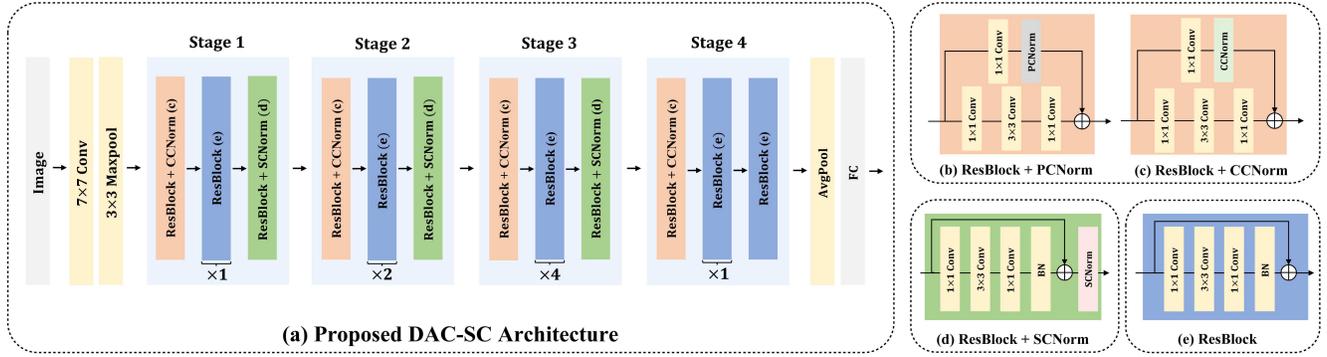}
  \caption{
  Overall architecture of DAC-SC. (a): DAC-SC is composed of ResNet50 with $CCNorm$ and $SCNorm$.
  (b), (c): The proposed $PCNorm$ and $CCNorm$ replace BN in the downsample layer, respectively.
  We use $CCNorm$ for DAC-SC and $PCNorm$ for DAC-P.
  (d): $SCNorm$ is attached at the end of stage1 to stage3.
  (e): A residual block is represented.
  }
  
  \label{fig:short-a}
\end{figure*}
In this context, we propose our $PCNorm$, depicted in Fig. \ref{fig:norms} (a).
$PCNorm$ is a normalization method that maintains an original content.
$PCNorm$ decomposes pre-and post-normalized features into their phase and amplitude, respectively.
Then, it combines $\rho$ with $\alpha^{norm}$. Finally, the composed frequency feature is transformed to spatial feature through inverse FT (IFT).

The $PCNorm$ is defined as follows:
\begin{equation}
PCNorm({f}) = IFT(compose(\alpha^{norm}, \rho)),
\label{eq:11}
\end{equation}
where $IFT$($\cdot$) denotes IFT. 
\subsection{Content Controlling Normalization (CCNorm)}
\label{sec:ccnorm}
$PCNorm$ prevents content from changing.
Further, we have a fundamental question regarding whether the variation of content is absolutely negative in DG.
We believe that the answer can be clarified by making the model learn itself to reduce the content change.
If the change is harmful to DG, the model would gradually decrease it during training.

As we explain in Sec. \ref{sec:3.2}, content change occurs due to $\mu$, which is the mean shift in normalization in the spatial domain. .
Thus, we adjust the degree of content change by introducing a learnable parameter $\lambda^c \in \mathbb{R}^{2}$. The content adjusting terms, ($\lambda^c_{norm}$, $\lambda^c_{org}$) $= softmax\left({\lambda^c}/{T_{c}}\right)$, where $T_{c}$ denotes the temperature value, determine the proportion of the normalized and original content, respectively.
Then, the content-adjusted feature $f^c$ is defined as follows:
\begin{equation}
    f^c = f - {\mu} \, \lambda^c_{norm}.
\label{eq:12}
\end{equation}

In the equation, $\lambda^c_{org}$ is omitted because it is a dummy variable for the performance, and we only consider the normalized content. 
If $\lambda^c_{norm}$ is 0, the phase of $f^{c}$ ($\rho^{c}$) is the same as $\rho$, and if $\lambda^c_{norm}$ is 1, $\rho^{c}$ is the same as $\rho^{norm}$.
Then, we propose the first advanced $PCNorm$, $CCNorm$, which employs $\rho^{c}$ instead of $\rho$ in $PCNorm$. The $CCNorm$ is defined as follows:
\begin{equation}
CCNorm({f}) = IFT(compose(\alpha^{norm}, \rho^{c})),
\label{eq:13}
\end{equation}
which is illustrated in Fig. \ref{fig:norms} (b).
The main idea of $CCNorm$ is to mitigate the content variation of normalization, not to entirely offset it.
Interestingly, $CCNorm$ performs better than $PCNorm$ in many experiments.
Sec. \ref{sec:5} discusses the effects of $CCNorm$ and insights from it.
\subsection{Style Controlling Normalization (SCNorm)}
\label{sec:scnorm}
It is common to remove style information through IN in DG\cite{Jin2020StyleNA,ibn}. 
Similar to the motivation of $CCNorm$, an essential question occurs regarding whether completely eliminating style is ideal for DG.
Thus, we identify the effect of adjusting the style elimination degree.

Therefore, we propose the other advanced version of $PCNorm$, $SCNorm$, that regulates the degree of style elimination. 
It is illustrated in Fig. \ref{fig:norms} (c).
The $SCNorm$ method mixes $\alpha$ with $\alpha^{norm}$ by applying learnable parameters. 
In $SCNorm$, the content is preserved in the same way as in $PCNorm$.
Analogous to $CCNorm$, the style-adjusting terms are $\lambda^s_{norm}$ and $\lambda^s_{org}$, which are the outputs of the learnable parameter, $\lambda^s \in \mathbb{R}^2$, obtained using the softmax. That is, ($\lambda^s_{norm}$, $\lambda^s_{org}$) $= softmax\left({\lambda^s}/{T_{s}}\right)$, where $T_{s}$ represents temperature value.  
The $SCNorm$ is formulated as follows:
\begin{equation}
\resizebox{1.\columnwidth}{!}{%
$SCNorm(f) = IFT(compose(\lambda^s_{norm} \, \alpha^{norm} + \lambda^s_{org} \, \alpha, \rho)).$%
}
\label{eq:14}
\end{equation}

Both adjusting terms, $\lambda^s_{norm}$ and $\lambda^s_{org}$, are learned to make the model independently determine the ratios of pre- and post-normalized style, respectively.
The IN is used by default to obtain $\alpha^{norm}$.
Similar to $\lambda^c$ in $CCNorm$, $SCNorm$ becomes an identity function when $\lambda^s_{norm}$ is 0, and if $\lambda^s_{norm}$ is 1, $SCNorm$ completely removes the style. Sec. \ref{sec:5} discusses its effectiveness.
\subsection{DAC-P and DAC-SC} 
\label{sec:4}
In this section, we introduce the ResNet-variant models, DAC-P, and DAC-SC.
DAC-P is the early model for feasibility, where $PCNorm$ is applied. DAC-SC is the primary model that adopts both $CCNorm$ and $SCNorm$. The overall architecture of DAC-SC is described in Fig. \ref{fig:short-a}.

In DAC-P, we replace BN in the downsample layer with $PCNorm$ (Fig. \ref{fig:short-a} (b)).
The downsample layer is included in the residual block, which is represented as $\mathcal{H}(x) + x$, where $x$ is the input feature, and $\mathcal{H}(\cdot)$ is the residual function.
Unlike other layers, the shape of $x$ in downsample layer changes to match that of $\mathcal{H}(x)$.
That is, the phase of $x$ inevitably changes, although for identity mapping it should be invariant.
Thus, the residual $\mathcal{H}(x)$ is approximated to the biased input whose content information is changed.
As it contains BN with the content variation problem, the content change in this layer becomes larger. 
Consequently, the biased approximation of $\mathcal{H}(x)$ degrades DG performance.

To relieve this, we substitute the BN in the downsample layer with $PCNorm$. 
It is to take advantage of the fact that $PCNorm$ preserves content.
First, the existing downsample layer of ResNet\cite{resnet}, $downsample(\cdot)$, is represented as follows:
\useshortskip
\begin{equation}
    downsample(x) = BatchNorm(Conv(x)),
\label{eq:15}
\end{equation}
where $BatchNorm(\cdot)$ indicates the BN, and $Conv(\cdot)$ is a 1$\times$1 convolution layer with a stride 2, and $x$ is the input feature.
On the other hand, the downsample layer of our DAC-P, $downsample_{p}(\cdot)$, is formulated as follows:
\begin{equation}
    downsample_{p}(x) = PCNorm(Conv(x)).
\label{eq:16}
\end{equation}

Next, we introduce the primary model, DAC-SC, where both $CCNorm$ and $SCNorm$ are applied.
As explained above, the content change in the downsample layer especially needs to be relieved.
In DAC-SC, traditional BN is replaced with $downsample_{c}$, which uses $CCNorm$ instead of $PCNorm$ (Fig. \ref{fig:short-a} (c)). Then,
$downsample_{c}$ is defined as follows:
\begin{equation}
    downsample_{c}(x) = CCNorm(Conv(x)).
\label{eq:17}
\end{equation}

In addition, DAC-SC also exploits $SCNorm$, which adjusts the degree of style elimination.
Previous studies\cite{Zhou2021DomainGW,Jin2020StyleNA} found that putting a style regularizer between residual blocks enhances performance.
Inspired by this, we inserted $SCNorm$ between residual blocks (Fig. \ref{fig:short-a} (d)).
Hence, the primary model, DAC-SC, determines the proper intensities of the content and style changes for DG.

\section{Experiment}
\label{sec:5}
\begin{table*}[]

\begin{center}
\adjustbox{max width=0.8\textwidth}{%
\begin{tabular}{lccccccccc}
\toprule
Model        & VLCS             & PACS             & Office-Home       & DomainNet        & TerraIncognita              & Avg              \\
\midrule
ERM\cite{Vapnik1998StatisticalLT}                       & 77.4 $\pm$ 0.3            & 85.7 $\pm$ 0.5            & 67.5 $\pm$ 0.5            & 41.2 $\pm$ 0.2            & 47.2 $\pm$ 0.4            & 63.8                       \\
IRM\cite{arjovsky2019invariant}                       & 78.5 $\pm$ 0.5            & 83.5 $\pm$ 0.8            & 64.3 $\pm$ 2.2            & 33.9 $\pm$ 2.8            & 47.6 $\pm$ 0.8            & 61.6                      \\
GroupDRO\cite{Sagawa2020DistributionallyRN}                  & 76.7 $\pm$ 0.6            & 84.4 $\pm$ 0.8            & 66.0 $\pm$ 0.7            & 33.3 $\pm$ 0.2            & 43.2 $\pm$ 1.1 & 60.7                      \\
Mixup\cite{Yan2020ImproveUD}                     & 77.4 $\pm$ 0.6            & 84.6 $\pm$ 0.6            & 68.1 $\pm$ 0.3            & 39.2 $\pm$ 0.1            & 47.9 $\pm$ 0.8 & 63.4                      \\
MLDG\cite{Li2018LearningTG}                      & 77.2 $\pm$ 0.4            & 84.9 $\pm$ 1.0            & 66.8 $\pm$ 0.6            & 41.2 $\pm$ 0.1            & 47.7 $\pm$ 0.9 & 63.6                      \\
CORAL\cite{sun2016deep}                     & \textbf{78.8 $\pm$ 0.6}            & 86.2 $\pm$ 0.3            & 68.7 $\pm$ 0.3            & 41.5 $\pm$ 0.1            & 47.6 $\pm$ 1.0 & 64.5                      \\
MMD\cite{Li2018DomainGW}                       & 77.5 $\pm$ 0.9            & 84.6 $\pm$ 0.5            & 66.3 $\pm$ 0.1            & 23.4 $\pm$ 9.5            & 42.2 $\pm$ 1.6 & 58.8                      \\
DANN\cite{Ganin2016DomainAdversarialTO}                      & 78.6 $\pm$ 0.4            & 83.6 $\pm$ 0.4            & 65.9 $\pm$ 0.6            & 38.3 $\pm$ 0.1            & 46.7 $\pm$ 0.5 & 62.6                      \\
CDANN\cite{Li2018DeepDG}                     & 77.5 $\pm$ 0.1            & 82.6 $\pm$ 0.9            & 65.8 $\pm$ 1.3            & 38.3 $\pm$ 0.3            & 45.8 $\pm$ 1.6 & 62.0                      \\
MTL\cite{Blanchard2021DomainGB}                       & 77.2 $\pm$ 0.4            & 84.6 $\pm$ 0.5            & 66.4 $\pm$ 0.5            & 40.6 $\pm$ 0.1            & 45.6 $\pm$ 1.2 & 62.9                      \\

ARM\cite{Zhang2021AdaptiveRM}                       & 77.6 $\pm$ 0.3            & 85.1 $\pm$ 0.4            & 64.8 $\pm$ 0.3            & 35.5 $\pm$ 0.2            & 45.5 $\pm$ 0.3 & 61.7                      \\
VREx\cite{Krueger2021OutofDistributionGV}                      & 78.3 $\pm$ 0.2            & 84.9 $\pm$ 0.6            & 66.4 $\pm$ 0.6            & 33.6 $\pm$ 2.9            & 46.4 $\pm$ 0.6 & 61.9                      \\
RSC\cite{Huang2020SelfChallengingIC}                     & 77.1 $\pm$ 0.5            & 85.2 $\pm$ 0.9            & 65.5 $\pm$ 0.9            & 38.9 $\pm$ 0.5            & 46.6 $\pm$ 1.0 & 62.7                      \\

IIB\cite{Li2022InvariantIB}                       & 77.2 $\pm$ 1.6            & 83.9 $\pm$ 0.2            & 68.6 $\pm$ 0.1            & 41.5 $\pm$ 2.3            & 45.8 $\pm$ 1.4 & 63.4                      \\
SelfReg\cite{kim2021selfreg}                   & 77.8 $\pm$ 0.9            & 85.6 $\pm$ 0.4            & 67.9 $\pm$ 0.7            & 42.8 $\pm$ 0.0            & 47.0 $\pm$ 0.3 & 64.2                      \\

SagNet\cite{Nam2021ReducingDG}                    & 77.8 $\pm$ 0.5            &  \underline{86.3 $\pm$ 0.2}            & 68.1 $\pm$ 0.1            & 40.3 $\pm$ 0.1            &  \underline{48.6 $\pm$ 1.0} & 64.2                      \\

\hline
\textbf{DAC-P (ours)}                  & 77.0 $\pm$ 0.6            & 85.6 $\pm$ 0.5            &  \underline{69.5 $\pm$ 0.1}            &  \underline{43.8 $\pm$ 0.3}            & \textbf{49.8 $\pm$ 0.2} &   \underline{65.1}                     \\
\textbf{DAC-SC (ours)}                 
&  \underline{78.7 $\pm$ 0.3}            & \textbf{87.5 $\pm$ 0.1}     & \textbf{70.3 $\pm$ 0.2}            & \textbf{44.9 $\pm$ 0.1}            & 46.5 $\pm$ 0.3 & \textbf{65.6}                      \\
\bottomrule

\end{tabular}}

\caption{Comparison with recent DG methods.
For the performances of IIB and SelfReg, we refer to each paper.
The other performance results are the reported numbers from DomainBed\cite{domainbed}. The best performance values are in bold and the second-best performance valuess are underlined.}
\label{tab:main}
\end{center}
\end{table*}
\subsection{Dataset}

\vspace*{-\baselineskip}
\begin{table}[H]
  \centering
  \renewcommand{\arraystretch}{0.75}

    \begin{tabular}{@{}lcccc@{}}
    \toprule
    \textbf{Dataset} & \textbf{Class} & \textbf{Image} & \textbf{Domain} \\
    \midrule
    VLCS\cite{pacs} & 7 & 9,991 & 4 \\
    PACS\cite{pacs} & 5 & 10,729 & 4 \\
    Office-Home\cite{office} & 65 & 15,588 & 4 \\
    DomainNet \cite{domain} & 345 & 586,575 & 6 \\
    TerraIncognita\cite{Beery2018RecognitionIT} & 10 & 24,788 & 4 \\
    \bottomrule
  \end{tabular}
  
  \caption{Description of five datasets, VLCS, PACS, Office-Home, DomainNet, and TerraIncognita. }
  \label{tab:data}
\end{table}
We evaluated the proposed methods on five DG benchmarks: VLCS, PACS, Office-Home, DomainNet, and TerraIncognita.
Table \ref{tab:data} summarizes the number of classes, images, and domains in each dataset.

\subsection{Experimental Details}
We chose ResNet50 as a backbone network, the same as the baseline model (ERM)\cite{Vapnik1998StatisticalLT}.
In DAC-P, the BN in all four downsample layers was replaced with $PCNorm$ layers.
For DAC-SC, four $CCNorm$ were inserted at the same locations as $PCNorm$ in DAC-P, and three $SCNorm$ were added at the ends of the first to third stages, respectively.
The affine transform layer of base normalization is transferred to the end of the proposed normalization layer.
The model was initialized with ImageNet\cite{deng2009imagenet} pre-trained weight. The elements of $\lambda^s$ and $\lambda^c$ were initialized with 0, and the temperatures $T_s$ and $T_c$ were set to 1e-1 and 1e-6, respectively.
For data augmentation, we randomly cropped images on a scale from 0.7 to 1.0 and resized them to 224×224 pixels. Then, we applied random horizontal flip, color jittering, and gray scaling. 
In training, we used a mini-batch size of 32, and the Nesterov SGD optimizer with a weight decay of 5e-4, a learning rate of 1e-4, and momentum of 0.9. 
For DomainNet dataset only, the learning rate of 1e-2 was applied.
We trained the proposed model for 20 epochs with 500 iterations each, except for DomainNet, which we trained with 7500 iterations and adopted a cosine annealing scheduler with early stopping (tolerance of 4).
All experiments were conducted four times and each performance was evaluated on the training-domain validation set, which reserved 20\% of source domain data.
The performance values were reported using the average performance for the entire domains, which was evaluated using a single out-of-training domain.
We conducted an exhaustive hyperparameter search for model selection and evaluated the models based on accuracy.
The hardware and software environments were Ubuntu 18.04, Python 3.8.13, PyTorch 1.12.1+cu113, CUDA 11.3, and a single NVIDIA A100 GPU.
\subsection{Comparison with SOTA Methods}
\label{sec:5.3}
The test accuracies of DAC-P, DAC-SC, and recent DG methods on five benchmarks are reported in Table \ref{tab:main}.
Specifically, DAC-SC, which exhibited the highest average performance improvement, achieved new SOTA results at 87.5$\%$, 70.3$\%$, and 44.9$\%$ on the PACS, Office-Home, and DomainNet benchmark datasets. 
These experimental results indicate that there can be proper degrees of content and style changes for DG. 
In the TerraIncognita dataset, DAC-P outperformed DAC-SC, which was expected due to the dataset characteristics. 
The TerraIncognita dataset consists of camera trap images in which objects are captured in four locations. The content between the domains is closely distributed, resulting in fewer domain gaps compared to other datasets. In these cases, preserving the content can be more effective than adjusting changes under a slight domain shift. From the results of this experiment, it is critical to preserve content in DG, but higher performance can be achieved when the preservation degree can be appropriately controlled. A more detailed discussion of this is presented in Sec. \ref{sec:cvc}.
\begin{table}
  \centering
  \resizebox{1.\columnwidth}{!}{
  \begin{tabular}{c|lccccccc}
    \toprule
    \multicolumn{2}{c}{Method} & P & VL & PA & OH & DN & TI & Avg \\
    \hline
    1 & ResNet & & 77.4 & 85.7 & 67.5 & 41.2 & 47.2 & 63.8\\
    2 & +$CCNorm$ & D & 77.1 & 86.0 & 69.9 & 44.7 & 48.1 & 65.1(+1.3)\\
    3 & \textbf{+$SCNorm$} & E & \textbf{78.7} & \textbf{87.4} & \textbf{70.3} & \textbf{44.9} & 46.5 &  \textbf{65.6(+0.5)}\\
    
     \hline
     4 & +$PCNorm$ & D & 77.0 & 85.6 & 69.5 & 43.8 & \textbf{49.8} & 65.1(+1.3) \\
     5 & +$SCNorm$ & E & 77.1 & 86.0 & 68.8 & 44.6 & 48.0 & 64.9(-0.2)\\
     \hline
     6 & +$SCNorm$ & D & 77.3 & 86.4 & 69.2 & 44.5 & 45.3 & 64.5(+0.7)\\
     7 & +$SCNorm$ & E & 77.3 & 86.8 & 69.2 & \textbf{44.9} & 43.0 & 64.3(-0.2)\\
     
     \hline  
     \end{tabular}}
        \caption{Ablation study on our methods. 
        VL, PA, OH, DN, and TI denote VLCS, PACS, Office-Home, DomainNet, and TerraIncognita, respectively.
        P indicates module position, and D and E of column P mean downsample layer and end of the stage in order.}
\label{tab:ab}
\end{table}
\subsection{Ablation Study}
In this section, we conducted various ablation studies focusing on the DAC-SC because it is the main model and adopts many advanced factors.

\textbf{Effect of CCNorm and SCNorm.}
To investigate the performance improvement effect with the two proposed modules, $CCNorm$ and $SCNorm$, we added them sequentially and individually to ResNet and compared their performance (Rows 2 and 3 of Table \ref{tab:ab}).
As described in Sec. \ref{sec:4}, $CCNorm$ is applied to the downsample layer and $SCNorm$ is inserted at the end of the stage. As the two modules are inserted in different positions, we added Column P, indicating the position of the applied module for clarity, where D and E denote the downsample layer and end of the stage, respectively.
An average performance improvement of 1.3\% compared to the baseline ResNet (ERM) occurs when only $CCNorm$ is added.
Furthermore, we added both $SCNorm$ and $CCNorm$, the DAC-SC, providing an extra improvement of 0.5$\%$ to the average performance.
It achieves the highest increase of 3.7$\%$ on the DomainNet dataset.

In contrast, applying $PCNorm$ and $SCNorm$ together displayed an average performance degradation of 0.2$\%$ compared to applying $PCNorm$ alone (Row 5). Considering the performance gains when $SCNorm$ is combined with $CCNorm$, it implies that $SCNorm$ works better with $CCNorm$ than $PCNorm$.
These experimental results demonstrate that there is an efficient combination of adjusted content and style, which DAC-SC can successfully  determine.

\textbf{Content Preserving vs. Content Controlling.}
\label{sec:cvc}
Rows 2 and 4 of Table \ref{tab:ab} reveal that $PCNorm$, which perfectly preserves content, and $CCNorm$, which controls content change, have the same average performance at 65.1$\%$. However, referring to the results of the individual datasets, $CCNorm$ outperforms $PCNorm$ on the other four datasets except TerraIncognita. 
It means that controlling content changes ($CCNorm$) is more beneficial in terms of generalization performance.

\textbf{Content Controlling vs. Style Controlling in Downsample Layer.}
We replace $CCNorm$ with $SCNorm$ in DAC-SC to verify that adjusting content changes is more effective than controlling style changes at the downsample layer.
Rows 6 and 7 of Table \ref{tab:ab} present the results when $SCNorm$ is inserted into the downsample layer. 
Compared to the baseline ResNet, applying $SCNorm$ to the downsample layer renders a slight average performance improvement of 0.7$\%$. 
However, this performance enhancement is low compared to the result of applying $CCNorm$ or $PCNorm$ to the downsample layer.
From these results, we infer that, although the adjustment to the style change in the downsample layer works lightly, the regulation of the content variation seems more appropriate in this layer.

\textbf{Which Normalization Fits on SCNorm?}
In $SCNorm$, we apply IN by default. 
In this experiment, We first compared the performances of our $SCNorm$ with those of the simple IN to clarify that $SCNorm$ is more effective than existing IN. Then, we observed the results when the IN in $SCNorm$ is replaced with the LN or BN to identify that it is optimal to select the IN compared to other methods.

As listed in Table \ref{tab:scn}, applying IN to $SCNorm$ is more effective for all datasets than using the simple IN (InNorm). Compared with the other two normalization methods (LN and BN) applied to $SCNorm$, the highest average performance is reached when IN is applied to $SCNorm$ except in the TerraIncognita dataset. In contrast, $SCNorm$ with LN consistently performs poorly compared with the other two normalization methods (IN and BN) except for the PACS dataset. These results indicate that it is optimal to apply IN to $SCNorm$.
\begin{table}
  \centering
  \renewcommand{\arraystretch}{0.75}
  \begin{tabular}{@{}lcccccc@{}}
    \toprule
    Method & VL & PA & OH & DN & TI & Avg \\
    \midrule
    InNorm & 76.0 & 87.2 & 65.6 & 44.0 & 42.7 &  63.1\\
    \midrule
    $SCNorm_{in}$ & \textbf{78.7} & \textbf{87.4} & \textbf{70.3} & \textbf{44.9} & 46.5 & \textbf{65.6}\\
    $SCNorm_{ln}$ & 75.0 & 87.3 & 67.9 & 44.3 & 44.3 & 63.8\\
    $SCNorm_{bn}$ & 77.0 & 86.6 & \textbf{70.3} & 44.8 & \textbf{46.9} & 65.1\\
    \bottomrule
  \end{tabular}
  \caption{Ablation study on base normalization in $SCNorm$.
  InNorm is the simple IN without $SCNorm$.
  $SCNorm_{in}$, $SCNorm_{ln}$, and $SCNorm_{bn}$ denote IN, LN, and BN applied to $SCNorm$, respectively.
  The abbreviations of the datasets are the same as in Table \ref{tab:ab}.}
  \label{tab:scn}
\end{table}
\begin{figure}[t]
  \centering
  \includegraphics[width=1.\linewidth]{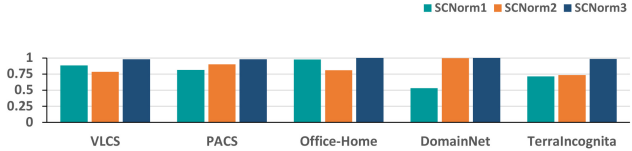}
  \caption{Comparison of the learned values of $\lambda^s_{org}$ in each $SCNorm$ layer.}
  \label{fig:viz}
\end{figure}

\textbf{Adjusting Style Elimination Degree.}
To identify the degree that style elimination is adjusted through $SCNorm$, we illustrate $\lambda^s_{org}$ in $SCNorm$ in Fig. \ref{fig:viz}, where
$\lambda^s_{org}$ is the weight of the original style, which determines the extent of the original style when it is mixed with the normalized style.
Moreover, $SCNorm$1, $SCNorm$2, and $SCNorm$3 indicate each $SCNorm$ module at the ends of the first, second, and third stages of the DAC-SC, respectively. 
As demonstrated, all $\lambda^s_{org}$ were generally distributed above 0.5, but no clear tendency was found for $\lambda^s_{org}$ of $SCNorm1$ and $SCNorm2$. However, all $\lambda^s_{org}$ in $SCNorm$3 had an average value of 0.9898, which is very close to 1, in all five datasets.
That is, the original style $\alpha$ is nearly preserved at the end of Stage 3.

\begin{figure}[t]
  \centering
   \includegraphics[width=1.\linewidth]{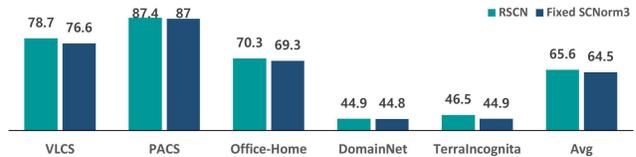}
   
   \caption{Accuracy comparison of DAC-SC with and without style controlling ($\lambda^s_{org}$).
   Fixed $SCNorm$3 indicates that $\lambda^s_{org}$ is set to a fixed number of 1.} 
   \label{fig:scnorm3}
\end{figure}

Then, we deduce that adjusting the style elimination is not required in $SCNorm$3 
because the reflection of the normalized style is negligible.
To clarify this, we set $\lambda^s_{org}$ in $SCNorm$3 to a fixed number at 1 and compared the performance.
As presented in Fig. \ref{fig:scnorm3}, the performance consistently drops when $\lambda^s_{org}$ is 1 in $SCNorm$3.
On the VLCS, Office-Home, and TerranIncognita benchmarks, the performance drops by more than 1.0\%.
This result confirms that even a small change in $\lambda^s_{org}$ significantly affects performance.
Hence, we infer that an adjustment is needed in the style elimination degree for DG, which $SCNorm$ successfully determines.

\section{Conclusion}
This paper addresses the content change problem in existing normalization methods in DG, and suggests an incipient approach that explores its improvements from the aspect of the frequency domain. For this, we provide a pioneering analysis of the quantitative content change through mathematical derivation. Based on this analysis, we propose novel normalization methods for DG, PCNorm, CCNorm, and SCNorm. 
With the proposed methods, the ResNet variant models, DAC-P and DAC-SC, both achieve SOTA performance on five DG benchmarks. The experimental results highlight the importance of content preservation in DG, and take a step further in variation adjustments in the existing normalization-based style extraction.

{\small
\bibliographystyle{ieee_fullname}
\bibliography{paper}
}

\end{document}